\documentclass[letterpaper]{article} % DO NOT CHANGE THIS
\usepackage{aaai2026}  % DO NOT CHANGE THIS
\usepackage{times}  % DO NOT CHANGE THIS
\usepackage{helvet}  % DO NOT CHANGE THIS
\usepackage{courier}  % DO NOT CHANGE THIS
\usepackage[hyphens]{url}  % DO NOT CHANGE THIS
\usepackage{graphicx} % DO NOT CHANGE THIS
\usepackage{natbib}  % DO NOT CHANGE THIS AND DO NOT ADD ANY OPTIONS TO IT
\usepackage{caption} % DO NOT CHANGE THIS AND DO NOT ADD ANY OPTIONS TO IT
\usepackage{subcaption}
\usepackage{algorithm}
\usepackage{algorithmic}
\usepackage{amsmath}
\usepackage{amssymb}
\usepackage{booktabs}
\usepackage{xcolor}
\definecolor{graytext}{gray}{0.5} % 定义半透明灰色文本颜色

\usepackage{multirow} 
\usepackage{newfloat}
\usepackage{listings}
\usepackage{color}
\definecolor{codegreen}{rgb}{0,0.6,0}
\definecolor{codegray}{rgb}{0.5,0.5,0.5}
\title{
    \raisebox{-0.1cm}{\includegraphics[width=1.0cm]{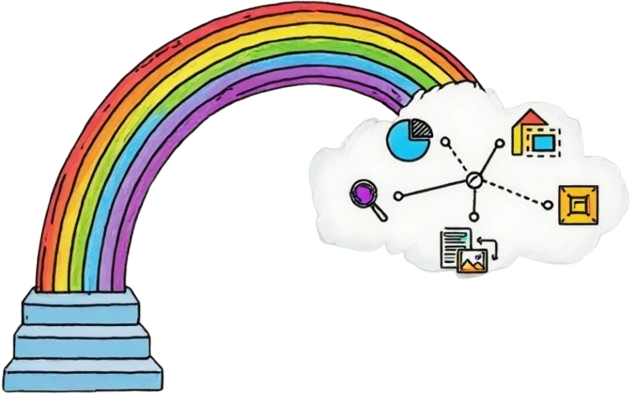}}\hspace{0.0cm}
    Visual Bridge: Universal Visual Perception Representations Generating
}
\author{
    Yilin Gao\textsuperscript{\rm 1 $\ast$}, Shuguang Dou\textsuperscript{\rm 2},
    Junzhou Li\textsuperscript{\rm 3},\\
    Zhiheng Yu\textsuperscript{\rm 2},
    Yin Li\textsuperscript{\rm 2},
    Dongsheng Jiang\textsuperscript{\rm 2},
    Shugong Xu\textsuperscript{\rm 4}
}

\affiliations{
    \textsuperscript{\rm 1}Shanghai University\\
    \textsuperscript{\rm 2}Huawei Technologies Co., Ltd.\\
    \textsuperscript{\rm 3}University of Science and Technology of China\\
    \textsuperscript{\rm 4}Xi'an Jiaotong-Liverpool University\\
    gaoyilin@shu.edu.cn,\,
    doushuguang@huawei.com,\,
    ljz0824@mail.ustc.edu.cn,\\
    yuzhiheng8@huawei.com,\,
    sherrylee90@163.com,\,
    dongsheng\_jiang@outlook.com,\,
    shugong.xu@xjtlu.edu.cn\\
    \textsuperscript{$\ast$}Work done during an internship at Huawei\\
    Both Shugong Xu and Dongsheng Jiang are corresponding authors.
}

\usepackage{bibentry}
\begin{document}

% \maketitle
\twocolumn[{
\renewcommand\twocolumn[1][]{#1}
\maketitle

\begin{center}
\label{fig:teaser}
\begin{minipage}[b]{0.65\linewidth}
  \centering
  \includegraphics[width=\textwidth]{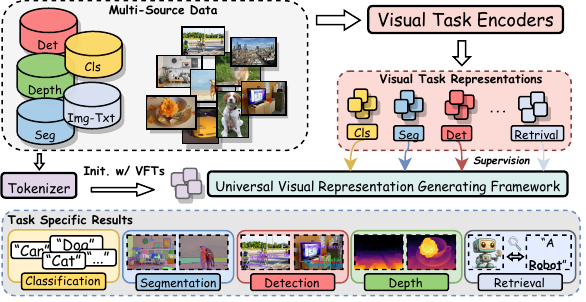}
  \par\medskip
  \small\textit{(a) One Universal Visual Representation Generating Framework for Multi-Tasks}
\end{minipage}
\hfill
\begin{minipage}[b]{0.3\linewidth}
  \centering
  \includegraphics[width=\textwidth]{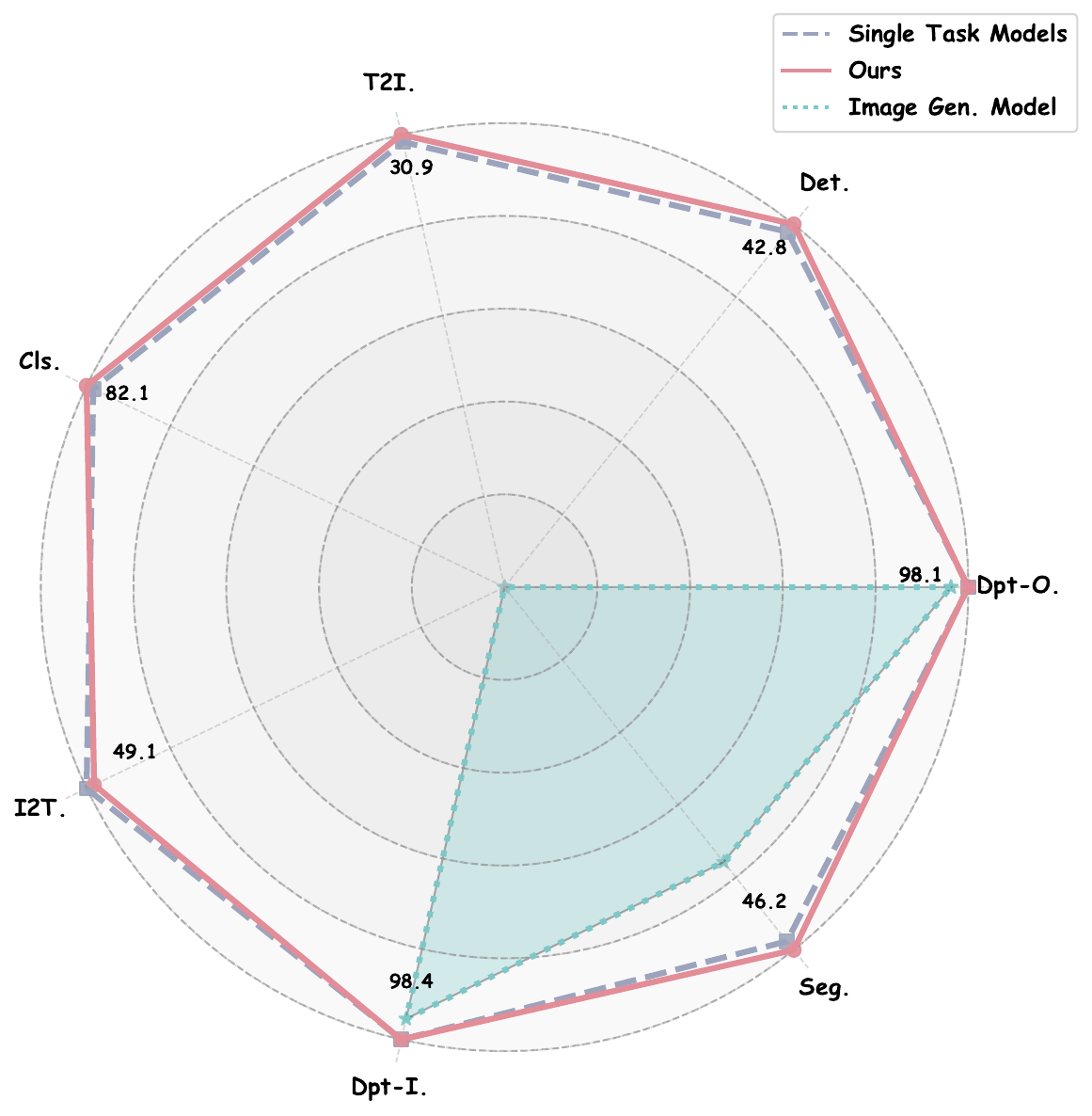}
  \par\medskip
  \small\textit{(b) Performance on Five Task Datasets}
\end{minipage}

\medskip
\parbox{1.0\linewidth}{%
% \small
\captionof{figure}{
    \textbf{(a)} We introduce \textit{Vision Bridge}, a universal framework that bridges image patch tokens and fundamental vision tasks through flow-based modeling. The architecture supports diverse downstream tasks, including classification, detection, segmentation, depth estimation, and image-text retrieval. \textit{Vision Bridge} enables task-agnostic representation learning and task-specific adaptation without introducing external data. 
    \textbf{(b)} Radar plot comparing performance across five core vision tasks. Our method consistently outperforms strong baselines on ImageNet-1K~\cite{deng2009imagenet}, COCO~\cite{coco}, ADE20K~\cite{ade20k}, and NYUv2~\cite{nyu}.
}
}

\end{center}
}]

\begin{abstract}
Recent advances in diffusion models have achieved remarkable success in isolated computer vision tasks such as text-to-image generation, depth estimation, and optical flow. However, these models are often restricted by a ``single-task-single-model'' paradigm, severely limiting their generalizability and scalability in multi-task scenarios. Motivated by the cross-domain generalization ability of large language models, we propose a universal visual perception framework based on \emph{flow matching} that can generate diverse visual representations across multiple tasks.
Our approach formulates the process as a universal flow-matching problem from image patch tokens to task-specific representations rather than an independent generation or regression problem. By leveraging a strong self-supervised foundation model as the anchor and introducing a multi-scale, circular task embedding mechanism, our method learns a universal velocity field to bridge the gap between heterogeneous tasks, supporting efficient and flexible representation transfer.
Extensive experiments on classification, detection, segmentation, depth estimation, and image-text retrieval demonstrate that our model achieves competitive performance in both zero-shot and fine-tuned settings, outperforming prior generalist and several specialist models. Ablation studies further validate the robustness, scalability, and generalization of our framework. Our work marks a significant step towards general-purpose visual perception, providing a solid foundation for future research in universal vision modeling.
\end{abstract}

\section{Introduction}
%扩散模型能否像LLM一样通用，适用于不同的CV任务
In recent years, diffusion models have achieved remarkable success in text-to-image generation, depth estimation, and optical flow estimation. However, these models are typically trained under a "single-task-single-model" paradigm, limiting their generalizability. In contrast, large-scale language models (LLMs) like GPT-4~\cite{gpt4} and DeepSeek-R1~\cite{ds2025r1} exhibit strong cross-domain zero-shot transfer capabilities, enabled by universal architectures and large-scale pretraining—demonstrating the clear advantages of general-purpose modeling. 
Motivated by this observation, the paper explores: \emph{Is it possible to achieve a universal generation of the diverse visual perception representation by diffusion models?}

%不同CV任务的范式壁垒
Generating representations for diverse visual perception tasks within a universal diffusion-based framework presents significant challenges. % 页数有点超了，就把model删了
% \begin{figure}[!htb]
% \centering
% \includegraphics[scale=0.55]{Images/Intro/intro.pdf}
% \caption{
%     \textbf{Comparison of existing paradigms for unifying visual representations.}
%     (a) \textbf{Latent-to-latent mapping}: Traditional methods transfer features across task-specific latent spaces, often requiring dedicated architectures for different tasks, which limits generalization and scalability.
%     (b) \textbf{Noise-to-image mapping}: Diffusion-based approaches unify perception via image generation, but discard non-generative tasks (e.g., classification, retrieval) and fail to handle the multi-scale and heterogeneous nature of diverse vision tasks.
%     (c) \textbf{Token-to-latent mapping} (\textbf{Ours}): Our framework aligns foundation model embeddings with task-specific latent representations through a universal flow, enabling flexible and generalizable visual perception across a wide range of tasks.
% }
% \label{fig:intro}
% \end{figure}
Classification models (e.g. ResNet~\cite{resnet}, ViT~\cite{vit}) are designed to encode global semantic features, detection models (e.g., Faster R-CNN~\cite{fasterr-cnn} ) depend on multi-scale feature pyramid networks (FPNs) to localize and recognize objects at various scales, and segmentation models (e.g., U-Net~\cite{unet}) rely on encoder–decoder architectures to recover fine-grained spatial details.
These structural differences highlight the difficulty of designing a single framework capable of effectively handling the heterogeneous requirements of multiple tasks. %同样，也是有点超了，就把vision删了

% Despite the remarkable success of diffusion models in isolated vision tasks, achieving \textbf{ universal visual perception} across diverse domains remains a fundamental challenge in the field of computer vision. Existing models are typically tailored to specific tasks and architectures, which hinders their generalizability and scalability when faced with the broad spectrum of real-world scenarios. This fragmentation significantly limits the flexibility and deployability of vision systems in multi-task environments.

Drawing inspiration from the cross-domain generalization capabilities exhibited by LLMs, we posit that a universal visual representation generating framework should possess the following essential characteristics:
\begin{itemize}
    \item \textbf{Task Generality}: The capacity to generate representations suitable for a variety of visual tasks without requiring separate models or extensive fine-tuning.
    \item \textbf{Architectural Flexibility}: The ability to adapt seamlessly to different network architectures and scales, supporting both global semantic reasoning and fine-grained spatial understanding.
    \item \textbf{Model Efficiency}: The ability to approximate the representations of larger models using compact architectures, enabling efficient deployment without significant performance loss.
\end{itemize}

To address these challenges, we propose a novel \textbf{universal flow-matching framework} for multi visual perception tasks. In contrast to previous works that treat perception tasks as an independent generation or regression problem~\cite{zhao2025diception, le2024diffusiongenerate} (detailed in the \textit{\textbf{\textbf{Appendix}}}), our approach formulates multi-vision perception tasks as \emph{flow matching problem from image patch tokens to task-specific representations}, as shown in Fig. 1. %Fig.~\ref{fig:teaser}. 很奇怪，交叉引用用不了
By leveraging powerful self-supervised models such as DINOv2~\cite{oquab2023dinov2} as the foundation and learning a universal velocity field conditioned on task embeddings, we enable smooth and efficient representation transfer across various vision tasks.

Our key contributions can be summarized as follows:
\begin{itemize}
    \item We introduce a general and scalable \emph{flow-matching paradigm} that unifies visual perception tasks by aligning foundation model tokens with task-specific representations, moving beyond traditional image-to-image translation or single-stage distillation methods.
    \item We design a multi-scale, circular task embedding scheme that enables flexible encoding and interpolation across heterogeneous vision tasks, supporting both dense and sparse prediction scenarios.
    % \item Extensive experiments across classification, detection, segmentation, depth estimation, and image-text retrieval demonstrate that our universal model achieves competitive performance in both zero-shot and fine-tuned settings, outperforming prior generalist models and several task-specific baselines.
    \item Extensive experiments of our model on five vision tasks demonstrate competitive performance in both zero-shot and fine-tuned settings, outperforming prior generalist models and task-specific baselines.
    % \item Extensive experiments of our model on classification, detection, segmentation, depth estimation, and image-text retrieval demonstrate the competitive performance in both zero-shot and fine-tuned settings, outperforming prior generalist models and task-specific baselines.
    % \item Comprehensive ablation studies validate the effectiveness of our sampling strategy and model capacity, highlighting the robustness and scalability of our framework.
\end{itemize}

In summary, our work takes an important step towards \textbf{general-purpose visual perception}, bridging the gap between task-specific and universal vision models. We believe our approach can serve as a solid foundation for future research in multi-task and foundation vision modeling.

\section{Related Work}
\begin{figure*}[htb]
\centering
\includegraphics[scale=1.1]{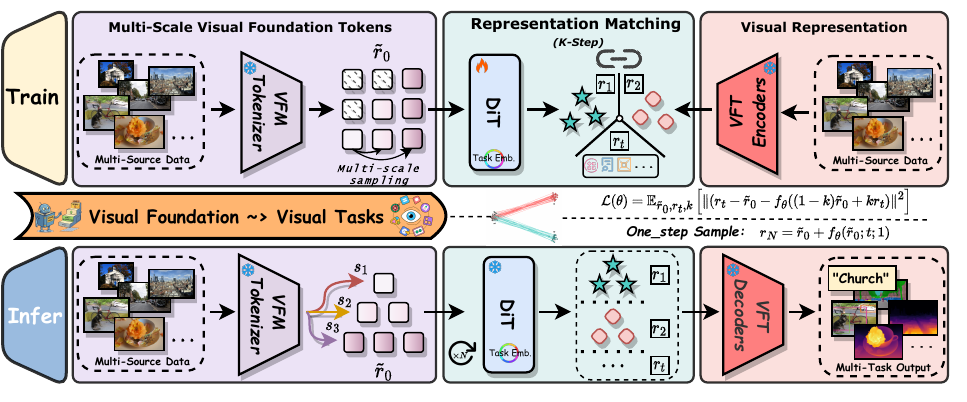}
\caption{
    \textbf{Overview of the proposed Visual Bridge.}
    During training, tokens from the foundation model are sampled and interpolated with task-specific representations at multiple scales. A universal velocity field, conditioned on circular task embeddings and learnable scale embeddings, models the dynamics at each step. During inference, the learned flow is integrated to generate task-specific outputs (e.g., bounding boxes, labels, segmentation masks) using dedicated decoders. The proposed architecture enables efficient and flexible unification of a wide range of visual perception tasks.
}
\label{fig:pipeline}
\end{figure*}

\subsection{Vision Foundation Models}
\label{sec:related_vfm}
Recent advances in vision foundation models (VFMs) have enabled powerful generalist representations through large-scale pre-training. Prominent examples include DINOv2~\cite{oquab2023dinov2}, CLIP~\cite{clip}, and MAE~\cite{mae}, which demonstrate strong cross-domain generalization capabilities. Task-specific architectures such as ViT~\cite{vit} for classification, DETR~\cite{detr} for detection, and U-Net~\cite{unet} or Mask2Former~\cite{mask2former} for segmentation achieve high performance on individual tasks but suffer from limited adaptability across modalities.

Efforts to transfer knowledge from VFMs to task-specific models~\cite{wang2024task,khanna2024explora,li2024matching,rajivc2023segment,zhong2024convolution, zhu2024unleashing} often require complex architectural adaptations (e.g., custom feature extractors) or fine-tuning strategies, which limit flexibility. While universal representation learning approaches~\cite{ren2024dino, he2022masked, oquab2023dinov2, caron2021emerging} aim to bridge this gap, they typically underperform compared to specialized models. For instance, methods like DINOv2~\cite{oquab2023dinov2} focus on self-supervised learning but struggle to match the task-specific accuracy of models trained on curated datasets.

Our work introduces a universal framework that bridges image patch tokens of VFMs to task-specific representations. By leveraging a sampling-based iterative design, it achieves consistent cross-task performance with minimal architectural changes. Unlike prior methods that prioritize either generality or task-specific accuracy, our approach enables flexible trade-offs between efficiency and precision, making it suitable for deployment in diverse scenarios.

\subsection{Multi-task Generalist Models}
\label{sec:related_multi}
Multi-task vision systems have primarily followed two paradigms: (1) encoder-decoder frameworks~\cite{lu2022unified,bachmann20244m}, where shared encoders map tasks into a common space; and (2) vision-language models (VLMs)~\cite{bai2023qwen, chen2024internvl, lu2024deepseek, ren2024pixellm, li2025llamavid}, which excel in multimodal understanding but struggle with general visual perception tasks. Recent works like ~\cite{lu2022unified, lu2024unified, mizrahi20234m, bachmann20244m} propose transformer-based architectures with task-specific decoders, but their reliance on large-scale, task-specific data limits practical deployment.

Diffusion-based approaches~\cite{le2024diffusiongenerate,zhao2025diception} further explore multi-task unification by reusing pretrained generative models. For example, One Diffusion~\cite{le2024diffusiongenerate} treats multi-task image generation as a shared latent structure problem, while DICEPTION~\cite{zhao2025diception} targets multi-task image understanding by reusing pretrained generative models. However, these methods are constrained by their dependency on image generation architectures, which prioritize reconstruction over task-specific representations. This leads to suboptimal performance in tasks like classification or retrieval.

In contrast, we formulate multi-vision perception tasks as a flow matching problem from Token-to-Representation, rather than Image-to-Image transformation. This approach: 
% (1) leverages the visual reasoning capabilities of VFMs; 
(1) establishes a task-agnostic representation bridge; and (2) enables seamless task switching with minimal architectural changes. Our sampling-based architecture further supports efficient deployment by balancing accuracy and cost, offering a scalable solution for real-world systems.

\section{Method}
\paragraph{Objective.} 
We formulate the process of multi-vision perception tasks as a \textit{universal flow matching} problem from image patch tokens to \textbf{Vision Foundation Task (VFT)} representations, as shown in Fig.~\ref{fig:pipeline}. Specifically, we treat established task-specific models (e.g., DeiT for classification, DETR for detection) as target representations and utilize powerful self-supervised vision models such as DINOv2~\cite{oquab2023dinov2} as the initial flow anchor, as opposed to VAE-based latent representations in generative models. This enables us to effectively exploit the intrinsic visual reasoning capabilities of vision foundation models, thereby advancing universal learning across diverse vision tasks.

Mathematically, let $\mathbf{x} \in \mathbb{R}^{H \times W \times C}$ denote a raw image sampled from a dataset $\mathcal{D} = \{\mathbf{x}_i\}_{i=1}^N$, where $N$ is the number of training samples. We extract image tokens $\mathbf{r}_0$ from a self-supervised VFM, such as DINOv2~\cite{oquab2023dinov2}, and task-specific representations $\mathbf{r}_t$ from corresponding encoders. Our goal is to learn a velocity field $f_\theta$ that maps $\mathbf{r}_0$ to $\mathbf{r}_t$ under a flow-matching objective, conditioned on task embeddings $e_t$, which captures the dynamics of task-specific transformations. This formulation enables efficient multi-task learning by aligning intermediate representations via a universal flow field.

\paragraph{Training as Universal Flow-matching.} 

\begin{algorithm}[!t]
\caption{Universal Flow-matching Approach}
\label{alg:reflow_train}
\begin{algorithmic}[1]
\REQUIRE 
Training dataset $\mathcal{D} = \{(x^{(i)})\}_{i=1}^N$, where $x^{(i)}$ represents an input image\\
Model parameters $\theta$, learning rate $\eta$, epochs $E$
\ENSURE 
Trained model parameters $\theta$
\STATE Randomly initialize $\theta$
\FOR{$e = 1$ to $E$}
    \FOR{each batch $x \sim \mathcal{D}$}
        \STATE $r_0 \leftarrow \text{DINOV2\_Tokenizer}(x)$ %\COMMENT{Extract initial tokens $r_0$ using the DINOV2 tokenizer}
        \STATE  $r_t \leftarrow E_{task}(x)$ %Extract $r_t$ using task-specific Encoder:
        \STATE Generate multi-scale tokens by sampling $r_0$ to various scales: $\tilde{\mathbf{r}}_0 = \text{MultiScalesample}(r_0)$.
        \STATE Sample $k \sim \text{Discrete}(0, K)$
        \STATE $r_k \leftarrow (1 - k/K) \tilde{\mathbf{r}}_0 + (k/K) r_t$
        \STATE Predict velocity field: $\hat{v}_k = f_\theta(r_k, k, s, e_t)$
        \STATE Compute ground-truth velocity: $v_k^{\text{true}} = r_t - \tilde{\mathbf{r}}_0$
        \STATE Compute loss: $\mathcal{L}(\theta) = \|\hat{v}_k - v_k^{\text{true}}\|^2$
        \STATE  $\theta \leftarrow \theta - \eta \nabla_\theta \mathcal{L}(\theta)$ %Update parameters via gradient descent:
    \ENDFOR
\ENDFOR
\end{algorithmic}
\end{algorithm}
Our training pipeline is based on a universal flow-matching paradigm, as detailed in Algorithm~\ref{alg:reflow_train}. At each training step, we sample a batch $\mathbf{x}$ from the dataset $\mathcal{D}$ and extract its VFM tokens $\mathbf{r}_0$ using a frozen DINOv2 tokenizer. Simultaneously, we extract task-specific representations $\mathbf{r}_t$ using a task-specific encoder $E_{task}$. In particular, to accommodate VFTs involving multi-scale features, we sample $r_0$ to match the resolutions present in $\mathbf{r}_t$, constructing multi-scale tokens $\tilde{\mathbf{r}}_0$ to prepare for subsequent multi-scale modeling.

These tokens are interpolated at discrete steps $k$ to generate intermediate states $\mathbf{r}_k$, which serve as inputs to the velocity field $f_\theta$. To ensure smooth transitions between $\tilde{\mathbf{r}}_0$ and $\mathbf{r}_t$, we employ a linear interpolation scheme:
\begin{equation}
    \mathbf{r}_k = \left(1 - \frac{k}{K}\right) \tilde{\mathbf{r}}_0 + \frac{k}{K} \mathbf{r}_t
\end{equation}
The velocity field $f_\theta$ predicts the dynamics of these interpolations, and the ground-truth velocity is computed as the difference between task and foundation representations:
\begin{equation}
    v_k^{\text{true}} = \mathbf{r}_t - \tilde{\mathbf{r}}_0
\end{equation}

To distinguish across different scales and tasks, we augment $\tilde{\mathbf{r}}_0$ with both scale and task embeddings. For scale embeddings, we introduce the learnable scale embedding module $\mathbf{s} \in \mathbb{R}^{L \times d}$, where $L$ denotes the number of pre-defined scale levels and $d$ is the embedding dimension. Each scale level corresponds to a specific resolution in the multi-scale token hierarchy. During training, the scale embedding $s$ is indexed according to the current token scale and concatenated with the input representation $\tilde{\mathbf{r}}_0$, enabling the model to condition its predictions on the spatial resolution. Formally, we define:
\begin{equation}
    s = \text{ScaleEmbed}(l),
\end{equation}
where $l \in \{0, ..., L-1\}$ indicates the current scale level. For task embeddings, instead of adopting the text-prefix-style used in T2I tasks~\cite{le2024diffusiongenerate}, we propose a \textit{circular task embedding} to map discrete task identities into continuous representations. Given $T$ tasks, each task index $t \in \{0, ..., T-1\}$ is mapped onto the unit circle by defining its angular position as:
\begin{equation}
    \theta_t = \frac{2\pi t}{T}.
\end{equation}
This angle is then expanded into a vector of frequencies and used to construct a $d$-dimensional embedding via sinusoidal functions:
\begin{equation}
    \begin{aligned}
    \mathbf{e}_t = \big[ & \cos(\theta_t),\ \sin(\theta_t), \\
                         & \cos(2\theta_t),\ \sin(2\theta_t), \\
                         & \dots, \\
                         & \cos\left(\tfrac{d}{2}\theta_t\right),\ \sin\left(\tfrac{d}{2}\theta_t\right) \big]^\top.
    \end{aligned}
\end{equation}
This design ensures that task representations reside on a smooth, periodic manifold, facilitating task variation and supporting multi-scale modeling within multi-scale vision foundation tasks.

The training objective aims to minimize the discrepancy between predicted and ground-truth velocities:
\begin{equation}
    \mathcal{L}(\theta) = \mathbb{E}\left[\|f_\theta(\mathbf{r}_k, k, s, e_t) - v_k^{\text{true}}\|^2\right]
\end{equation}
This flow-matching objective encourages $f_\theta$ to capture task-specific dynamics while preserving the semantic structure of the foundation model.

\paragraph{Inference as Flow Integration.} 
During inference, we employ a deterministic integration process to evolve $\mathbf{r}_0$ into $\mathbf{r}_t$ for a given task $t$. Starting from the VFM representation $\mathbf{r}_0$, we sample $r_0$ to generate multi-scale tokens $\tilde{\mathbf{r}}_0$ and iteratively update the state using the learned velocity field $f_\theta$. The integration is implemented using Euler discretization:
\begin{equation}
    r_{n+1} \leftarrow r_{n} + f_\theta(r_n, n, s, e_t) / N
\end{equation}
where $N$ denotes the total step size and $n$ represents the current sampling steps. The final state $\mathbf{r}_N$ is decoded into the target output (e.g., bounding boxes, class labels) through a task-specific decoder $D_{task}$. This integration procedure is detailed in Algorithm~\ref{alg:octoflow_inference}.

\begin{algorithm}[htbp]
\caption{Inference Algorithm (Euler Method)}
\label{alg:octoflow_inference}
\begin{algorithmic}[1]
\REQUIRE 
Input image $x$, target task $t$, number of integration steps $N$
\ENSURE 
Sampled output $y \in \mathcal{Y}$ (e.g., bounding box, class label)
\STATE $r_0 \leftarrow \text{DINOV2\_Tokenizer}(x)$ %Extract $r_0$ using DINOV2 Tokenizer:
\STATE Generate multi-scale tokens by sampling $r_0$ to different scales: $\tilde{\mathbf{r}}_0 = \text{MultiScalesample}(r_0)$.
% \STATE $\Delta k = \frac{K}{N}$
\STATE Initialize state: $r \leftarrow \tilde{\mathbf{r}}_0$
\FOR{$n = 0$ to $N - 1$}
    % \STATE $k = n \cdot \Delta k$
    \STATE Predict velocity at current state: $v = f_\theta(r_n, n, s, e_t)$
    % \STATE Update state: $r_{n+1} \leftarrow r_{n} + \Delta k \cdot v$
    \STATE Update state using Euler's method: $r_{n+1} \leftarrow r_n + v/N$
\ENDFOR
% \STATE Decode final state: $y = \text{Task\_Decoder}(r_{N})$
\STATE  $y = D_{task}(r_N)$ %Decode final state into task-specific output:
\RETURN $y$
\end{algorithmic}
\end{algorithm}

% \begin{lstlisting}[language=Python, caption={Circular Task Embedding Layer Implementation}, label=lst:circular_task_embedding]
% import torch
% import math

% class CircularTaskEmbedding(torch.nn.Module):
%     def __init__(self, num_tasks, d_model=64):
%         super().__init__()
%         assert d_model % 2 == 0, "d_model must be even"
%         self.num_tasks = num_tasks
%         self.d_model = d_model
        
%         # Generate circular angles evenly spaced in [0, 2pi)
%         angles = torch.linspace(0, 2 * math.pi, num_tasks + 1)[:-1]  # (num_tasks, )
        
%         # Expand to (num_tasks, d_model // 2)
%         freqs = angles.unsqueeze(1).expand(-1, d_model // 2)
        
%         # Compute embeddings using cosine and sine basis
%         embedding = torch.cat([
%             torch.cos(freqs),
%             torch.sin(freqs)
%         ], dim=1)  # (num_tasks, d_model)

%         self.register_buffer('embedding', embedding)

%     def forward(self, task_ids):
%         """
%         Args:
%             task_ids (LongTensor): shape (batch_size, )
%         Returns:
%             Tensor: shape (batch_size, d_model)
%         """
%         return self.embedding[task_ids]
% \end{lstlisting}

\section{Experiments}
\subsection{Implementation Details}
\paragraph{Data.}
We evaluate our method on diverse vision tasks: image classification, object detection, semantic segmentation, depth estimation, and image-text retrieval. We use standard benchmarks without introducing external data: \textbf{ImageNet-1K} for classification, \textbf{COCO} for detection and retrieval, \textbf{ADE20K} for segmentation, and \textbf{NYUv2}/\textbf{KITTI} for indoor/outdoor depth estimation. All datasets are used in their standard splits, consistent with recent foundation model evaluations, following conventions in recent multi-task learning and vision foundation model studies.

\paragraph{Training.}
Our framework is built on the \textbf{DINOv2-base} vision foundation model~\cite{oquab2023dinov2}. Task-specific heads are based on standard architectures: DeiT (ViT-B)~\cite{deit}, DETR~\cite{detr}, Mask2Former~\cite{mask2former}, CLIP~\cite{clip}, and Depth Anything~\cite{depth_anything}. The diffusion component uses \textbf{DiT-B}~\cite{dit}.
We use a detection-style data pipeline: images in a batch are padded to the maximum resolution to preserve aspect ratios. Models are trained for 300 epochs with AdamW ($1 \times 10^{-4}$ learning rate, 0.01 weight decay) and a cosine scheduler.
For fair evaluation of representation, we keep backbone fixed during fine-tuning and only update task heads. Ablation studies explore different DiT scales to assess model capacity.

\subsection{Results on Visual Perception Tasks}
\paragraph{Classification.}
We evaluate the quality of visual representations on ImageNet-1K~\cite{deng2009imagenet} using DeiT-B~\cite{deit} as the target architecture. Our goal is to assess whether our framework learns transferable features that match or exceed those from task-specific pre-training.
\begin{table}[!t]
\centering
\begin{tabular}{c|c|c}
\toprule
Method & Acc@1 (\%) &  Patch Size\\
\midrule
DeiT-B Baseline & \textbf{81.8} & 14$\times$14 \\
\midrule
MAE~\cite{mae} & 79.8 & 14$\times$14 \\
CLIP~\cite{clip} & 78.1 & 7$\times$7 \\
OSD & 81.2 & 16$\times$16 \\
Noise$\dagger$ & 0.1 & 16$\times$16 \\
\midrule
\textcolor{graytext}{Ours (Zero-shot)} & \underline{\textcolor{graytext}{81.5}} & \textcolor{graytext}{16$\times$16} \\
Ours (Fine-tuned) & \textbf{82.1} & 16$\times$16 \\
\bottomrule
\end{tabular}
\caption{\textbf{Zero-shot and fine-tuned classification performance (Top-1 Accuracy \%) on ImageNet-1K}. ($\dagger$) denotes using random noise as input with DINOV2 embedding as a condition.}
\label{tab:classification}
\end{table}
As shown in Table~\ref{tab:classification}, our method achieves a zero-shot accuracy of \textbf{81.5\%}, closely matching the supervised baseline (81.8\%). Fine-tuning only the classification head further improves performance to \textbf{82.1\%}, demonstrating strong representation quality and adaptability.

An intuitive attempt is to fully follow the pipeline of the flow-matching model, which uses random noise as $r_0$ and the embedding extracted by the vision foundation models as the condition. However, this results in poor performance (0.1\%), likely due to the lack of meaningful initialization and insufficient conditioning for fine-grained semantics.

To better understand the impact of different vision foundation models, we compare our approach against MAE~\cite{mae} and CLIP~\cite{clip}. When transferred to DeiT-B without fine-tuning, MAE obtains 79.8\%, while CLIP yields a significantly lower accuracy of 78.1\%. The performance gap may be due to CLIP’s extremely low-resolution representation: it uses a 32× downsampling strategy (from 224×224 input to 7×7 patch tokens), which severely limits its ability to capture fine-grained visual details.
In contrast, our method, built upon DINOv2~\cite{oquab2023dinov2}, benefits from a dense and high-resolution tokenization scheme (with a patch size of 16×16) and self-distillation during training, resulting in richer and more localized visual representations. These advantages enable our model to outperform both MAE and CLIP under identical downstream settings, validating the effectiveness of DINOv2 as a strong foundation for multi-task learning. See \textbf{\textit{Appendix}} for further analysis of \textbf{Task Emb.} \& \textbf{Tokenizer}.

We also compare our method with a One-Step Distillation (OSD) variant under the same network architecture and training setup. This alternative strategy achieves an accuracy of 81.2\%, which is slightly lower than ours. This suggests that generative learning approach still retains strong representational power and outperforms distillation-based methods that rely on a single-stage knowledge transfer.

% Comparing with MAE~\cite{mae} and CLIP~\cite{clip}, we find that both underperform when transferred to DeiT-B without fine-tuning, achieving 79.8\% and 78.1\%, respectively. CLIP’s low resolution (7×7 patches) severely limits its discriminative power, while MAE lacks the strong semantic alignment of our approach.

% Our method benefits from DINOv2~\cite{oquab2023dinov2}, which provides dense, high-resolution tokens (16×16 patch size) and benefits from self-distillation. These properties yield richer, more localized representations, enabling our method to outperform both MAE and CLIP under the same downstream setup.

% We further compare with One-step Distillation (OSD) under identical settings. While OSD achieves 81.2\%, our generative approach shows superior performance, suggesting that diffusion-based modeling retains strong representational power and outperforms single-stage distillation strategies.

\paragraph{Detection.}
We evaluate our method on the object detection task using the COCO dataset~\cite{coco}, with DETR~\cite{detr} as the target architecture equipped with a ResNet-50 backbone.

Our approach demonstrates strong zero-shot transfer capability, achieving an mAP of \textbf{39.2} without any task-specific adaptation. This result indicates that the representations learned by our foundation model can also capture rich semantic and spatial structures, enabling effective deployment to downstream architectures even in the absence of fine-tuning. When fine-tuning only the detection head, we further improve the performance to \textbf{42.8}, surpassing the original DETR baseline by a clear margin.

Notably, this cross-backbone compatibility highlights the \textbf{universality and robustness} of our learned representations and framework, making our approach a promising foundation for multi-task and multi-architecture vision systems.

\begin{table}[!t]
\centering

\begin{tabular}{c|c|c}
\toprule
Method & mAP (\%) & Encoder \\
\midrule
DETR Baseline & \textbf{41.9} & ResNet-50 \\
\midrule
\textcolor{graytext}{Ours (Zero-shot)} & \textcolor{graytext}{39.2} & \textcolor{graytext}{DiT-B} \\
Ours (Fine-tuned) & \textbf{42.8} & DiT-B \\
\bottomrule
\end{tabular}
\caption{Comparison of object detection performance (mAP \%) on COCO.}
\label{tab:detection}
\end{table}

\paragraph{Segmentation.}
We evaluate our framework on the ADE20K~\cite{zhou2017scene}, comparing our \textit{DiT-B} against a ResNet50-based \textit{Mask2Former}~\cite{mask2former} with \textbf{multi-scales}. As shown in Table~\ref{tab:ade20k}, our method achieves competitive performance in a zero-shot manner, attaining 37.9 PQ, 24.5 AP, and 44.6 mIoU. Although the performance is slightly insufficient, our method obtains a more fine-grained segmentation result. See \textit{\textbf{Appendix}} for details.

Furthermore, with fine-tuning, our method demonstrates clear improvements across all metrics, confirming its strong adaptability to downstream tasks. 

\begin{table}[!t]
\centering
\begin{tabular}{c|ccc}
\toprule
Method & PQ & AP & mIoU \\
\midrule
Mask2Former (ResNet50) & 39.7 & \textbf{26.5} & 46.1 \\
\midrule
\textcolor{graytext}{Ours (Zero-shot)} & \textcolor{graytext}{37.9} & \textcolor{graytext}{24.5} & \textcolor{graytext}{44.6} \\
Ours (Fine-tuned) & \textbf{40.0} & 26.4 & \textbf{46.2} \\
\bottomrule
\end{tabular}
\caption{Semantic segmentation performance on ADE20K (PQ / AP / mIoU).}
\label{tab:ade20k}
\end{table}

\paragraph{Image-Text Retrieval.}
We further evaluate the cross-modal alignment capability of our framework on the image-text retrieval task using the COCO dataset~\cite{coco}. Following standard evaluation protocols, we report Recall at K (R@K) for K=1, 5, and 10 in both text-to-image (T2I) and image-to-text (I2T) settings.

Our method demonstrates strong zero-shot transfer performance when compared to CLIP-B~\cite{clip}, a widely recognized foundation model for vision-language learning. As shown in Table~\ref{tab:retrieval}, our approach achieves T2I scores of \textbf{30.9} (R@1), \textbf{56.2} (R@5), and \textbf{67.4} (R@10), outperforming CLIP-B across all metrics. In the I2T direction, our results are highly competitive, with scores of 49.1 (R@1), 74.9 (R@5), and 83.6 (R@10), closely matching those of CLIP-B.

% The observed gains in T2I retrieval suggest that our model learns more discriminative and semantically aligned textual representations conditioned on visual content, which is critical for accurate cross-modal matching from language queries. The comparable performance in I2T indicates that our visual representations remain robust and well-aligned with language, despite not being explicitly trained under contrastive learning as in CLIP.

% Moreover, unlike CLIP, which operates at a coarse patch resolution of 7×7 due to its 32× downsampling strategy, our method benefits from a higher-resolution tokenization scheme (16×16 patches). This enables finer-grained modeling of visual content, likely contributing to the improved performance in text-driven retrieval tasks.

\begin{table}[!t]
\centering
\setlength{\tabcolsep}{1.5mm} % 减小列间距（默认是6pt）
\begin{tabular}{c|ccc|ccc}
\toprule
\multirow{2}{*}{Method} & \multicolumn{3}{c|}{Text-to-Image (T2I)} & \multicolumn{3}{c}{Image-to-Text (I2T)} \\
& R@1 & R@5 & R@10 & R@1 & R@5 & R@10 \\
\midrule
% CLIP-B~\cite{clip} & 30.35 & 54.78 & 66.09 & 50.00 & 74.98 & 83.22 \\
% Ours & \textbf{30.88} & \textbf{56.15} & \textbf{67.40} & 49.12 & 74.88 & 83.64 \\
CLIP-B & 30.4 & 54.8 & 66.1 & \textbf{50.0} & \textbf{74.9} & 83.2 \\
Ours & \textbf{30.9} & \textbf{56.2} & \textbf{67.4} & 49.1 & \textbf{74.9} & \textbf{83.6} \\
\bottomrule
\end{tabular}
\caption{Zero-shot image-text retrieval performance (Recall@K, \%) on COCO.}
\label{tab:retrieval}
\end{table}

\paragraph{Depth.}
We evaluate our framework on monocular depth estimation using two standard benchmarks: KITTI~\cite{kitti} and NYUv2~\cite{nyu}. Following common evaluation protocols, we report two widely used metrics: Absolute Relative Error (AbsRel $\downarrow$) and the percentage of predicted depths(i.e., $\delta_1$ $\uparrow$).

As shown in Table~\ref{tab:depth_kitti_nyu}, our method achieves strong performance across both datasets. On KITTI, we attain an AbsRel of 0.048 and $\delta_1$ of 0.981, closely approaching the performance of Depth Anything~\cite{yang2024depth} and Depth Anything v2~\cite{yang2024depth2}, outperforming Depth Pro~\cite{bochkovskii2024depth}, which are trained explicitly for depth estimation. On NYUv2, we achieve an AbsRel of 0.056 and $\delta_1$ of 0.984, demonstrating high-quality metric depth estimation. See \textit{\textbf{Appendix}} for qualitative analysis.

Our model is trained to mimic the output of Depth Anything in-domain through a learned flow-based mapping, allowing us to inherit its strong geometric priors while enabling efficient deployment.
% without reliance on task-specific training. This further validates the effectiveness of distilling depth-aware representations from powerful task-specific models into generalist frameworks.
Compared to other universal or generalist vision models — such as Painter~\cite{wang2023images}, Unified-IO~\cite{lu2022unified}, and DICEPTION~\cite{zhao2025diception} — our method shows significant improvements. For instance, we outperform OneDiffusion~\cite{le2024diffusiongenerate} by 8.0\% in $\delta_1$ on KITTI and improve over DICEPTION by 4.8\% in the same metric. These results demonstrate that our framework better preserves spatial and geometric structure in learned representations, enabling superior performance in dense prediction tasks like depth estimation.

\begin{table}[!t]
\centering
\small
\setlength{\tabcolsep}{1mm} % 减小列间距（默认是6pt）
\begin{tabular}{@{}c|cc|cc@{}}
\toprule
\multirow{2}{*}{Method} & \multicolumn{2}{c|}{KITTI} & \multicolumn{2}{c}{NYUv2} \\
\cline{2-5}
 & AbsRel$\downarrow$ & $\delta_1$$\uparrow$ & AbsRel$\downarrow$ & $\delta_1$$\uparrow$ \\
\midrule
MiDaS~\cite{midas}   & 0.236  & 0.630 & 0.111 & 0.885 \\
Omnidata~\cite{omnidata}  & 0.149  & 0.835 & 0.074 & 0.945 \\
Metric3D v2~\cite{hu2024metric3d}   & 0.052  & 0.979 & 0.039 & 0.979 \\
DiverseDepth~\cite{diversedepth}  & 0.190  & 0.704 & 0.117 & 0.875 \\
LeReS~\cite{leres}  & 0.149  & 0.784 & 0.090 & 0.916 \\
HDN~\cite{hdn}  & 0.115  & 0.867 & 0.069 & 0.948 \\
GeoWizard~\cite{fu2024geowizard}  & 0.097  & 0.921 & 0.052 & 0.966 \\
DepthFM~\cite{gui2024depthfm}  & 0.083  & 0.934 & 0.065 & 0.956 \\
Marigold~\cite{ke2024repurposing}  & 0.099  & 0.916 & \underline{0.055} & 0.964 \\
GeoWizard~\cite{fu2024geowizard}  & 0.129  & 0.851 & 0.059 & 0.959 \\
DepthFM~\cite{gui2024depthfm}  & 0.174  & 0.718 & 0.082 & 0.932 \\
Genpercept~\cite{xu2024diffusion}  & 0.094  & 0.923 & 0.091 & 0.932 \\
Depth Pro  & 0.055  & 0.974 & \textbf{0.042} & \underline{0.977} \\
DepthAnything  & 0.046  & 0.982 & 0.056 & \textbf{0.984} \\
DepthAnything v2  & \textbf{0.045}  & \textbf{0.983} & 0.056 & \textbf{0.984} \\
\midrule
4M-XL~\cite{mizrahi20234m}  & 0.105 & 0.896 & 0.068 & 0.951 \\
Painter  & 0.324  & 0.393 & 0.046 & 0.979 \\
Unified-IO  & 0.188  & 0.699 & 0.059 & 0.970 \\
OneDiffusion & 0.101  & 0.908 & 0.087 & 0.924 \\
DICEPTION  & 0.075  & 0.945 & 0.072 & 0.939 \\
\midrule
Ours  & \underline{0.048} & \underline{0.981} & 0.056 & \textbf{0.984} \\
\bottomrule
\end{tabular}
\caption{Quantitative comparison of depth estimation on KITTI and NYUv2 datasets.}
\label{tab:depth_kitti_nyu}
\end{table}

\subsection{Ablation Study}
\paragraph{Ablation on Flow Sampling}

% To investigate how sample steps of flow model affects task performance, we conduct an ablation study using DiT-S~\cite{dit}. As shown in Table~\ref{tab:ablation_flow}, increasing the number of sampling steps leads to consistent improvements in classification accuracy on ImageNet~\cite{deng2009imagenet}.

% With only a single forward pass through the flow model (1-Flow), our method achieves 77.3\% top-1 accuracy. This already demonstrates strong mimicry capability despite not being explicitly trained for classification. When increasing the number of sampling steps to two (2-Flow), the accuracy improves to 77.78\%, narrowing the gap to the original DeiT-S baseline (79.8\%). Further increasing to ten sampling steps (10-Flow) yields 77.93\%, showing that more refined iterative decoding can better recover the target representation.

% This trend indicates that our framework benefits from additional sampling steps in the flow model, which enables more accurate reconstruction of the target feature space. Importantly, this is achieved without direct fine-tuning or access to the training data — highlighting the expressive power of diffusion-based framework for visual perception.

We evaluate how the number of flow sampling steps affects classification accuracy using DiT-S~\cite{dit} on ImageNet~\cite{deng2009imagenet}. As shown in Table~\ref{tab:ablation_flow}, increasing the number of steps leads to consistent performance gains.

With a single forward pass (1-Flow), our method achieves 77.3\% top-1 accuracy, showing strong mimicry of DeiT-S without explicit classification training. Using two steps (2-Flow), the accuracy improves to 77.8\%, narrowing the gap to the baseline (79.8\%). With ten steps (10-Flow), we reach 77.9\%, demonstrating that iterative refinement better recovers the target representation.

This indicates that more sampling steps enhance feature reconstruction through progressive updates, without requiring fine-tuning or access to training data — highlighting the expressive power of our flow-based framework.

\begin{table}[!t]
\centering
\begin{tabular}{c|c|c}
\toprule
Method & Acc@1 (\%) & Flow Steps \\
\midrule
DeiT-S (Target) & 79.8 & - \\
\midrule
\multirow{3}{*}{Ours} & 77.3 & 1 \\
& 77.8 & 2 \\
& \textbf{77.9} & 10 \\
\bottomrule
\end{tabular}
\caption{Ablation study on the number of flow sampling steps. Top-1 accuracy (\%) on ImageNet using DiT-S as the student model to simulate DeiT-S.}
\label{tab:ablation_flow}
\end{table}

\paragraph{Ablation on Model Capacity}
% To evaluate the scalability of our framework in emulating stronger vision transformers, we conduct experiments where a lightweight \textit{DiT-S} is used to simulate significantly larger models: \textit{DeiT-B} and \textit{DeiT-H}. As shown in Table~\ref{tab:capacity_ablation}, despite its architectural simplicity, our method achieves competitive performance in a zero-shot manner — 79.46\% and 83.92\% Top-1 accuracy for DeiT-B and DeiT-H targets respectively, outperforming existing One-step distillation.

% Importantly, with only a single flow step and no access to training data or teacher gradients, our method already shows strong alignment with complex target models. After fine-tuning, our approach further improves to 81.78\% on DeiT-B, surpassing both its zero-shot result and the performance of One-step distillation. These results demonstrate that our framework enables more effective knowledge transfer than single-step mimicry, even when using a much smaller backbone.

To assess our framework’s ability to emulate stronger vision transformers, we use a lightweight \textit{DiT-S} to simulate larger models (\textit{DeiT-B}, \textit{DeiT-H}), as shown in Table~\ref{tab:capacity_ablation}. Despite its simplicity, our method achieves competitive zero-shot accuracy — 79.46\% and 83.92\% Top-1 for DeiT-B and DeiT-H targets — outperforming existing one-step distillation (OSD).

Notably, without access to training data or teacher gradients, our method aligns well with complex targets using only a single flow step. Fine-tuning further improves performance to 81.78\% on DeiT-B. This demonstrates that our approach enables effective knowledge, even with a smaller backbone.

\begin{table}[!t]
\centering
\begin{tabular}{c|c|c}
\toprule
Method & Acc@1 (\%) & Target Model  \\
\midrule
DeiT-B & 81.8 & -  \\
DeiT-H & 85.2 & -  \\
\midrule
% OSD (Zero-shot) & - & DeiT-B  \\
OSD (Zero-shot) & \underline{83.79} & DeiT-H  \\
\midrule
Ours (Zero-shot) & 79.46 & DeiT-B  \\
Ours (Fine-tuned) & \textbf{81.78} & DeiT-B  \\
Ours (Zero-shot) & \textbf{83.97} & DeiT-H  \\
\bottomrule
\end{tabular}
\caption{Comparison of zero-shot and fine-tuned classification performance (Top-1 Accuracy \%) on ImageNet. DiT-S is used to simulate increasingly powerful targets.}
\label{tab:capacity_ablation}
\end{table}

\subsection{Visualizations and Analysis}
\begin{figure}[!htbp]
\centering
\includegraphics[width=0.9\linewidth]{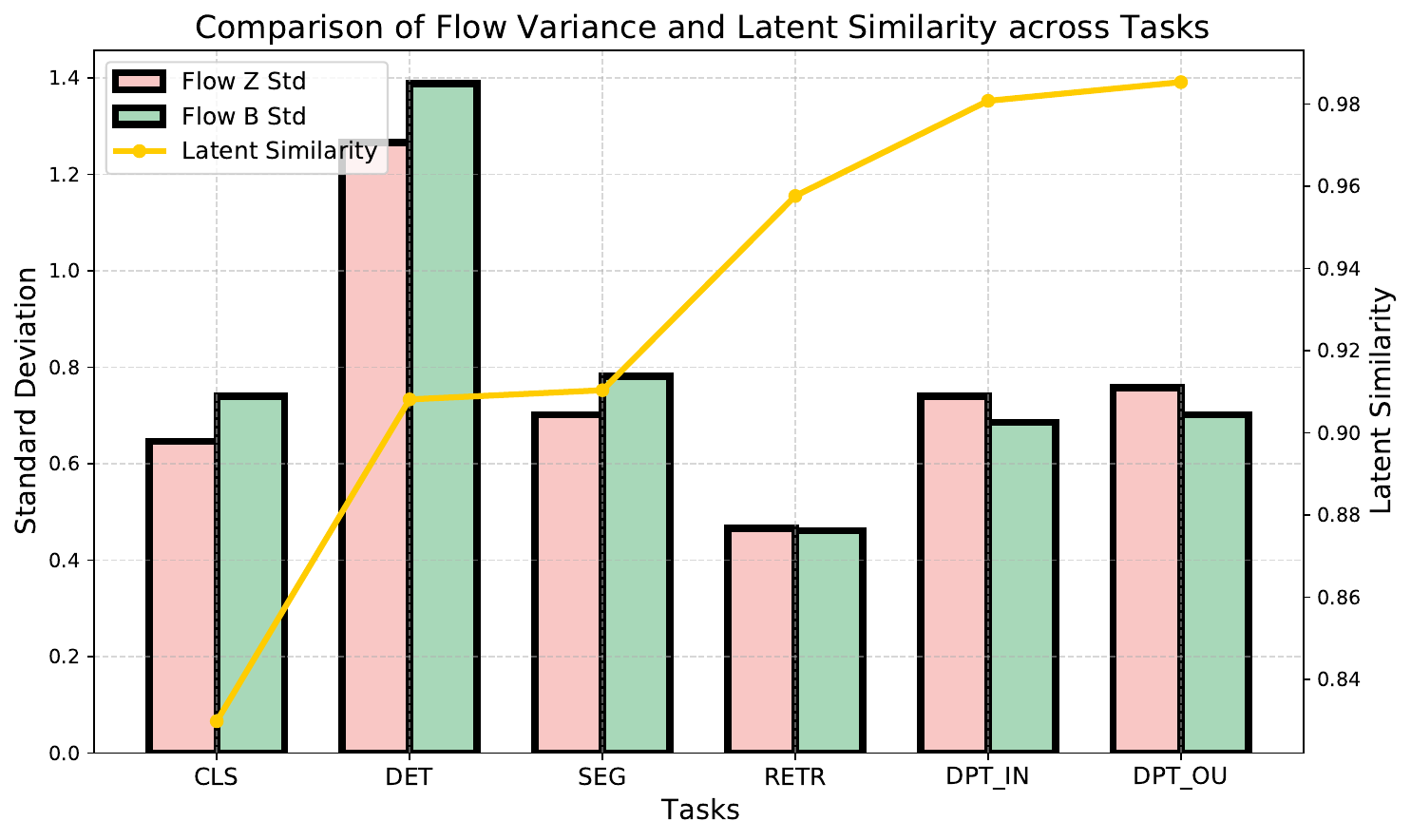}
\caption{
Similarity (line plot) and variance (bar plot) of generated (\textit{Flow z}) and target (\textit{Flow b}) latents across different models. Each bar represents the average standard deviation of features per dimension; the line denotes cosine similarity between \textit{Flow z} and \textit{Flow b}.
}
\label{fig:sim_var}
\end{figure}

\paragraph{Latent Similarity and Variance}
To better understand the relationship between generated and target latents, we analyze their feature similarity (measured by cosine similarity) and distribution variance across samples. As shown in Fig.~\ref{fig:sim_var}, higher similarity between \textit{Flow Z} (generated latent) and \textit{Flow B} (target latent) generally correlates with improved zero-shot performance.

Interestingly, we observe that models with lower latent variance often exhibit worse zero-shot accuracy, even when similarity is high. We hypothesize that flow-matching favors global feature consistency over preserving rare or extreme patterns, resulting in more compact latent distributions. While this may limit expressiveness in the zero-shot setting, it encourages robust feature learning that adapts well under fine-tuning — consistent with our observation of improved performance after adaptation.

\paragraph{Feature Evolution Analysis}
\begin{figure}[!t]
    \centering
    \centering
    \includegraphics[width=0.9\linewidth]{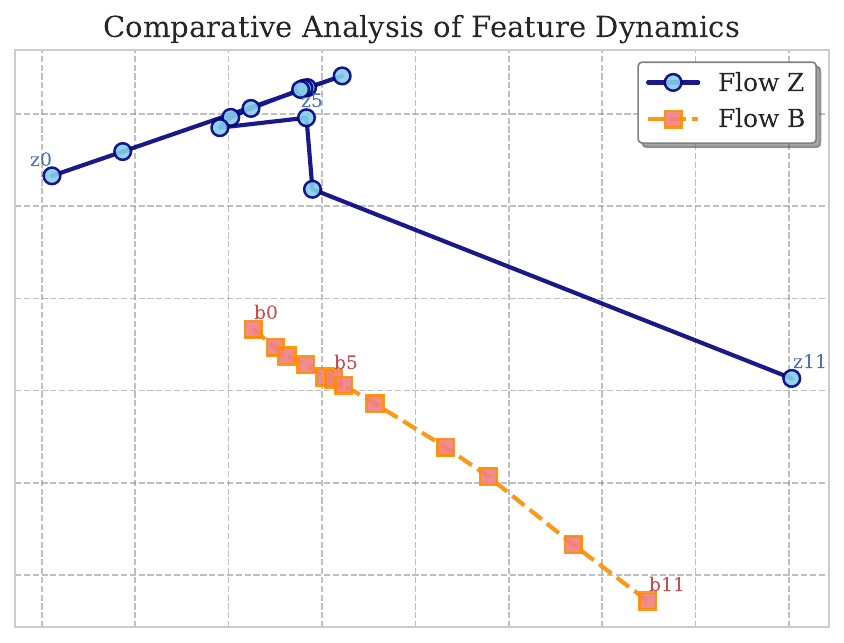}
    \caption{
        \textbf{Comparative Analysis of Feature Evolution Dynamics.} 
        Flow Z demonstrates an exploratory strategy characterized by early-stage feature exploration followed by rapid convergence. 
        Flow B exhibits a progressive refinement pattern with incremental adjustments toward the target. 
    }
    \label{fig:evolution}
\end{figure}
We analyze feature evolution using PCA visualization as shown in Fig.~\ref{fig:evolution}. The generative Flow Z exhibits a two-phase pattern: initial exploration of the feature space through iterations, followed by sharp convergence after locating the target region. In contrast, the supervised Flow B maintains steady incremental adjustments throughout, resulting in smoother but slower progress. The distinct trajectories reveal fundamental differences in optimization strategies—generative methods prioritize global exploration while supervised approaches focus on local refinement. These observations highlight the strengths of different learning paradigms in representation learning.
% To quantitatively analyze the dynamic behavior of different learning paradigms, we visualize the evolution of feature representations through principal component analysis (PCA) reduction. As shown in Figure~\ref{fig:evolution}, the proposed generative approach (Flow Z) demonstrates a distinct evolutionary pattern compared to the supervised baseline (Flow B).

% Flow Z initially explores the feature space through multiple noisy iterations, akin to an exploratory phase where the model identifies potential solution manifolds. Once the target region is located (typically after 5-7 steps), the trajectory undergoes rapid convergence toward the final representation. This behavior mirrors the cognitive process of hypothesis generation followed by targeted refinement.

% In contrast, the supervised method (Flow B) adopts a more conservative strategy. Its trajectory progresses through incremental adjustments with consistent step sizes, maintaining gradual improvements throughout the entire process. This iterative refinement mechanism results in smoother but slower convergence, reflecting the inherent constraints of label-guided optimization.

% The spatial separation between trajectories (controlled via coordinate offsets in visualization) further highlights the fundamental difference in optimization dynamics: generative methods prioritize global exploration while supervised approaches focus on local refinement. These findings provide empirical evidence supporting the complementary nature of different learning paradigms in representation learning.
\section{Discussion and Conclusion}
We present a universal visual perception generating framework based on flow matching, enabling a single framework to generate task-specific representations across diverse vision tasks. Our method achieves strong performance in both zero-shot and fine-tuned settings on classification, detection, segmentation, depth estimation, and image-text retrieval — demonstrating the effectiveness and versatility of the proposed flow-matching paradigm for general-purpose visual representation.

Despite these promising results, our current approach relies on task-specific encoders for supervision during training, and scaling to a large number of tasks remains non-trivial. In future work, we aim to explore fully end-to-end training to reduce dependency on external models, extend our framework to open-world and continual learning scenarios, and investigate universal modeling across modalities beyond vision.
In conclusion, our work advances the pursuit of general-purpose vision models and provides a foundation for future research in scalable and universal visual perception systems.

% Uncomment the following to link to your code, datasets, an extended version or similar.
% You must keep this block between (not within) the abstract and the main body of the paper.
% \begin{links}
%     \link{Code}{https://aaai.org/example/code}
%     \link{Datasets}{https://aaai.org/example/datasets}
%     \link{Extended version}{https://aaai.org/example/extended-version}
% \end{links}
    
\bibliography{aaai2026}

@article{ade20k,
  title={Semantic understanding of scenes through the ade20k dataset},
  author={Zhou, Bolei and Zhao, Hang and Puig, Xavier and Xiao, Tete and Fidler, Sanja and Barriuso, Adela and Torralba, Antonio},
  journal={International Journal of Computer Vision},
  volume={127},
  pages={302--321},
  year={2019},
  publisher={Springer}
}

@article{kitti,
  title={Vision meets robotics: The kitti dataset},
  author={Geiger, Andreas and Lenz, Philip and Stiller, Christoph and Urtasun, Raquel},
  journal={The International Journal of Robotics Research},
  volume={32},
  number={11},
  pages={1231--1237},
  year={2013},
  publisher={Sage Publications Sage UK: London, England}
}

@inproceedings{nyu,
  author    = {Nathan Silberman, Derek Hoiem, Pushmeet Kohli and Rob Fergus},
  title     = {Indoor Segmentation and Support Inference from RGBD Images},
  booktitle = {ECCV},
  year      = {2012}
}

@inproceedings{yang2024depth,
  title={Depth anything: Unleashing the power of large-scale unlabeled data},
  author={Yang, Lihe and Kang, Bingyi and Huang, Zilong and Xu, Xiaogang and Feng, Jiashi and Zhao, Hengshuang},
  booktitle={Proceedings of the IEEE/CVF Conference on Computer Vision and Pattern Recognition},
  pages={10371--10381},
  year={2024}
}

@article{yang2024depth2,
  title={Depth Anything V2},
  author={Yang, Lihe and Kang, Bingyi and Huang, Zilong and Zhao, Zhen and Xu, Xiaogang and Feng, Jiashi and Zhao, Hengshuang},
  journal={arXiv preprint arXiv:2406.09414},
  year={2024}
}

@inproceedings{leres,
  title={Learning to recover 3d scene shape from a single image},
  author={Yin, Wei and Zhang, Jianming and Wang, Oliver and Niklaus, Simon and Mai, Long and Chen, Simon and Shen, Chunhua},
  booktitle={Proceedings of the IEEE/CVF Conference on Computer Vision and Pattern Recognition},
  pages={204--213},
  year={2021}
}

@article{hu2024metric3d,
  title={Metric3D v2: A Versatile Monocular Geometric Foundation Model for Zero-shot Metric Depth and Surface Normal Estimation},
  author={Hu, Mu and Yin, Wei and Zhang, Chi and Cai, Zhipeng and Long, Xiaoxiao and Chen, Hao and Wang, Kaixuan and Yu, Gang and Shen, Chunhua and Shen, Shaojie},
  journal={arXiv preprint arXiv:2404.15506},
  year={2024}
}

@inproceedings{hdn,
  title={Multi-view Aggregation Network for Dichotomous Image Segmentation},
  author={Yu, Qian and Zhao, Xiaoqi and Pang, Youwei and Zhang, Lihe and Lu, Huchuan},
  booktitle={Proceedings of the IEEE/CVF Conference on Computer Vision and Pattern Recognition},
  pages={3921--3930},
  year={2024}
}

@article{diversedepth,
  title={Diversedepth: Affine-invariant depth prediction using diverse data},
  author={Yin, Wei and Wang, Xinlong and Shen, Chunhua and Liu, Yifan and Tian, Zhi and Xu, Songcen and Sun, Changming and Renyin, Dou},
  journal={arXiv preprint arXiv:2002.00569},
  year={2020}
}

@inproceedings{omnidata,
  title={Omnidata: A scalable pipeline for making multi-task mid-level vision datasets from 3d scans},
  author={Eftekhar, Ainaz and Sax, Alexander and Malik, Jitendra and Zamir, Amir},
  booktitle={Proceedings of the IEEE/CVF International Conference on Computer Vision},
  pages={10786--10796},
  year={2021}
}

@article{midas,
  title={Towards robust monocular depth estimation: Mixing datasets for zero-shot cross-dataset transfer},
  author={Ranftl, Ren{\'e} and Lasinger, Katrin and Hafner, David and Schindler, Konrad and Koltun, Vladlen},
  journal={IEEE transactions on pattern analysis and machine intelligence},
  volume={44},
  number={3},
  pages={1623--1637},
  year={2020},
  publisher={IEEE}
}

@article{bochkovskii2024depth,
  title={Depth pro: Sharp monocular metric depth in less than a second},
  author={Bochkovskii, Aleksei and Delaunoy, Ama{\"e}l and Germain, Hugo and Santos, Marcel and Zhou, Yichao and Richter, Stephan R and Koltun, Vladlen},
  journal={arXiv preprint arXiv:2410.02073},
  year={2024}
}

@inproceedings{he2022masked,
  title={Masked autoencoders are scalable vision learners},
  author={He, Kaiming and Chen, Xinlei and Xie, Saining and Li, Yanghao and Doll{\'a}r, Piotr and Girshick, Ross},
  booktitle={Proceedings of the IEEE/CVF conference on computer vision and pattern recognition},
  pages={16000--16009},
  year={2022}
}

@inproceedings{caron2021emerging,
  title={Emerging properties in self-supervised vision transformers},
  author={Caron, Mathilde and Touvron, Hugo and Misra, Ishan and J{\'e}gou, Herv{\'e} and Mairal, Julien and Bojanowski, Piotr and Joulin, Armand},
  booktitle={Proceedings of the IEEE/CVF international conference on computer vision},
  pages={9650--9660},
  year={2021}
}

@inproceedings{wang2023images,
  title={Images speak in images: A generalist painter for in-context visual learning},
  author={Wang, Xinlong and Wang, Wen and Cao, Yue and Shen, Chunhua and Huang, Tiejun},
  booktitle={Proceedings of the IEEE/CVF Conference on Computer Vision and Pattern Recognition},
  pages={6830--6839},
  year={2023}
}

@inproceedings{lu2022unified,
  title={Unified-io: A unified model for vision, language, and multi-modal tasks},
  author={Lu, Jiasen and Clark, Christopher and Zellers, Rowan and Mottaghi, Roozbeh and Kembhavi, Aniruddha},
  booktitle={The Eleventh International Conference on Learning Representations},
  year={2022}
}

@inproceedings{fu2024geowizard,
  title={Geowizard: Unleashing the diffusion priors for 3d geometry estimation from a single image},
  author={Fu, Xiao and Yin, Wei and Hu, Mu and Wang, Kaixuan and Ma, Yuexin and Tan, Ping and Shen, Shaojie and Lin, Dahua and Long, Xiaoxiao},
  booktitle={European Conference on Computer Vision},
  pages={241--258},
  year={2024},
  organization={Springer}
}

@article{gui2024depthfm,
  title={Depthfm: Fast monocular depth estimation with flow matching},
  author={Gui, Ming and Schusterbauer, Johannes and Prestel, Ulrich and Ma, Pingchuan and Kotovenko, Dmytro and Grebenkova, Olga and Baumann, Stefan Andreas and Hu, Vincent Tao and Ommer, Bj{\"o}rn},
  journal={arXiv preprint arXiv:2403.13788},
  year={2024}
}

@inproceedings{lu2024unified,
  title={Unified-IO 2: Scaling Autoregressive Multimodal Models with Vision Language Audio and Action},
  author={Lu, Jiasen and Clark, Christopher and Lee, Sangho and Zhang, Zichen and Khosla, Savya and Marten, Ryan and Hoiem, Derek and Kembhavi, Aniruddha},
  booktitle={Proceedings of the IEEE/CVF Conference on Computer Vision and Pattern Recognition},
  pages={26439--26455},
  year={2024}
}

@article{mizrahi20234m,
  title={4m: Massively multimodal masked modeling},
  author={Mizrahi, David and Bachmann, Roman and Kar, Oguzhan and Yeo, Teresa and Gao, Mingfei and Dehghan, Afshin and Zamir, Amir},
  journal={Advances in Neural Information Processing Systems},
  volume={36},
  pages={58363--58408},
  year={2023}
}

@article{bachmann20244m,
  title={4M-21: An Any-to-Any Vision Model for Tens of Tasks and Modalities},
  author={Bachmann, Roman and Kar, O{\u{g}}uzhan Fatih and Mizrahi, David and Garjani, Ali and Gao, Mingfei and Griffiths, David and Hu, Jiaming and Dehghan, Afshin and Zamir, Amir},
  journal={arXiv preprint arXiv:2406.09406},
  year={2024}
}

@article{xu2024diffusion,
  title={Diffusion models trained with large data are transferable visual models},
  author={Xu, Guangkai and Ge, Yongtao and Liu, Mingyu and Fan, Chengxiang and Xie, Kangyang and Zhao, Zhiyue and Chen, Hao and Shen, Chunhua},
  journal={arXiv preprint arXiv:2403.06090},
  year={2024}
}

@inproceedings{ke2024repurposing,
  title={Repurposing diffusion-based image generators for monocular depth estimation},
  author={Ke, Bingxin and Obukhov, Anton and Huang, Shengyu and Metzger, Nando and Daudt, Rodrigo Caye and Schindler, Konrad},
  booktitle={Proceedings of the IEEE/CVF Conference on Computer Vision and Pattern Recognition},
  pages={9492--9502},
  year={2024}
}

@article{ren2024dino,
  title={DINO-X: A Unified Vision Model for Open-World Object Detection and Understanding},
  author={Ren, Tianhe and Chen, Yihao and Jiang, Qing and Zeng, Zhaoyang and Xiong, Yuda and Liu, Wenlong and Ma, Zhengyu and Shen, Junyi and Gao, Yuan and Jiang, Xiaoke and others},
  journal={arXiv preprint arXiv:2411.14347},
  year={2024}
}

@article{wang2024task,
  title={Task-Aware Low-Rank Adaptation of Segment Anything Model},
  author={Wang, Xuehao and Ye, Feiyang and Zhang, Yu},
  journal={arXiv preprint arXiv:2403.10971},
  year={2024}
}

@article{khanna2024explora,
  title={ExPLoRA: Parameter-Efficient Extended Pre-Training to Adapt Vision Transformers under Domain Shifts},
  author={Khanna, Samar and Irgau, Medhanie and Lobell, David B and Ermon, Stefano},
  journal={arXiv preprint arXiv:2406.10973},
  year={2024}
}

@inproceedings{li2024matching,
  title={Matching Anything by Segmenting Anything},
  author={Li, Siyuan and Ke, Lei and Danelljan, Martin and Piccinelli, Luigi and Segu, Mattia and Van Gool, Luc and Yu, Fisher},
  booktitle={Proceedings of the IEEE/CVF Conference on Computer Vision and Pattern Recognition},
  pages={18963--18973},
  year={2024}
}

@article{rajivc2023segment,
  title={Segment anything meets point tracking},
  author={Raji{\v{c}}, Frano and Ke, Lei and Tai, Yu-Wing and Tang, Chi-Keung and Danelljan, Martin and Yu, Fisher},
  journal={arXiv preprint arXiv:2307.01197},
  year={2023}
}

@article{zhong2024convolution,
  title={Convolution meets lora: Parameter efficient finetuning for segment anything model},
  author={Zhong, Zihan and Tang, Zhiqiang and He, Tong and Fang, Haoyang and Yuan, Chun},
  journal={arXiv preprint arXiv:2401.17868},
  year={2024}
}

@article{zhu2024unleashing,
  title={Unleashing the potential of the diffusion model in few-shot semantic segmentation},
  author={Zhu, Muzhi and Liu, Yang and Luo, Zekai and Jing, Chenchen and Chen, Hao and Xu, Guangkai and Wang, Xinlong and Shen, Chunhua},
  journal={arXiv preprint arXiv:2410.02369},
  year={2024}
}

@misc{le2024diffusiongenerate,
      title={One Diffusion to Generate Them All}, 
      author={Duong H. Le and Tuan Pham and Sangho Lee and Christopher Clark and Aniruddha Kembhavi and Stephan Mandt and Ranjay Krishna and Jiasen Lu},
      year={2024},
      eprint={2411.16318},
      archivePrefix={arXiv},
      primaryClass={cs.CV},
      url={https://arxiv.org/abs/2411.16318}, 
}

@article{bai2023qwen,
  title={Qwen-vl: A frontier large vision-language model with versatile abilities},
  author={Bai, Jinze and Bai, Shuai and Yang, Shusheng and Wang, Shijie and Tan, Sinan and Wang, Peng and Lin, Junyang and Zhou, Chang and Zhou, Jingren},
  journal={arXiv preprint arXiv:2308.12966},
  year={2023}
}

@inproceedings{chen2024internvl,
  title={Internvl: Scaling up vision foundation models and aligning for generic visual-linguistic tasks},
  author={Chen, Zhe and Wu, Jiannan and Wang, Wenhai and Su, Weijie and Chen, Guo and Xing, Sen and Zhong, Muyan and Zhang, Qinglong and Zhu, Xizhou and Lu, Lewei and others},
  booktitle={Proceedings of the IEEE/CVF Conference on Computer Vision and Pattern Recognition},
  pages={24185--24198},
  year={2024}
}

@article{lu2024deepseek,
  title={Deepseek-vl: towards real-world vision-language understanding},
  author={Lu, Haoyu and Liu, Wen and Zhang, Bo and Wang, Bingxuan and Dong, Kai and Liu, Bo and Sun, Jingxiang and Ren, Tongzheng and Li, Zhuoshu and Yang, Hao and others},
  journal={arXiv preprint arXiv:2403.05525},
  year={2024}
}

@inproceedings{ren2024pixellm,
  title={Pixellm: Pixel reasoning with large multimodal model},
  author={Ren, Zhongwei and Huang, Zhicheng and Wei, Yunchao and Zhao, Yao and Fu, Dongmei and Feng, Jiashi and Jin, Xiaojie},
  booktitle={Proceedings of the IEEE/CVF Conference on Computer Vision and Pattern Recognition},
  pages={26374--26383},
  year={2024}
}

@inproceedings{li2025llamavid,
  title={Llama-vid: An image is worth 2 tokens in large language models},
  author={Li, Yanwei and Wang, Chengyao and Jia, Jiaya},
  booktitle={European Conference on Computer Vision},
  pages={323--340},
  year={2025},
  organization={Springer}
}

@article{lipman2022flow,
  title={Flow matching for generative modeling},
  author={Lipman, Yaron and Chen, Ricky TQ and Ben-Hamu, Heli and Nickel, Maximilian and Le, Matt},
  journal={arXiv preprint arXiv:2210.02747},
  year={2022}
}

@article{oquab2023dinov2,
  title={Dinov2: Learning robust visual features without supervision},
  author={Oquab, Maxime and Darcet, Timoth{\'e}e and Moutakanni, Th{\'e}o and Vo, Huy and Szafraniec, Marc and Khalidov, Vasil and Fernandez, Pierre and Haziza, Daniel and Massa, Francisco and El-Nouby, Alaaeldin and others},
  journal={arXiv preprint arXiv:2304.07193},
  year={2023}
}

@inproceedings{unet,
  title={U-net: Convolutional networks for biomedical image segmentation},
  author={Ronneberger, Olaf and Fischer, Philipp and Brox, Thomas},
  booktitle={International Conference on Medical image computing and computer-assisted intervention},
  pages={234--241},
  year={2015},
  organization={Springer}
}

@article{fasterr-cnn,
  title={Faster R-CNN: Towards real-time object detection with region proposal networks},
  author={Ren, Shaoqing and He, Kaiming and Girshick, Ross and Sun, Jian},
  journal={IEEE transactions on pattern analysis and machine intelligence},
  volume={39},
  number={6},
  pages={1137--1149},
  year={2016},
  publisher={IEEE}
}

@inproceedings{resnet,
  title={Deep residual learning for image recognition},
  author={He, Kaiming and Zhang, Xiangyu and Ren, Shaoqing and Sun, Jian},
  booktitle={Proceedings of the IEEE conference on computer vision and pattern recognition},
  pages={770--778},
  year={2016}
}

@article{vit,
  title={An image is worth 16x16 words: Transformers for image recognition at scale},
  author={Dosovitskiy, Alexey and Beyer, Lucas and Kolesnikov, Alexander and Weissenborn, Dirk and Zhai, Xiaohua and Unterthiner, Thomas and Dehghani, Mostafa and Minderer, Matthias and Heigold, Georg and Gelly, Sylvain and others},
  journal={arXiv preprint arXiv:2010.11929},
  year={2020}
}

@inproceedings{deit,
  title={Training data-efficient image transformers \& distillation through attention},
  author={Touvron, Hugo and Cord, Matthieu and Douze, Matthijs and Massa, Francisco and Sablayrolles, Alexandre and J{\'e}gou, Herv{\'e}},
  booktitle={International conference on machine learning},
  pages={10347--10357},
  year={2021},
  organization={PMLR}
}

@inproceedings{detr,
  title={End-to-end object detection with transformers},
  author={Carion, Nicolas and Massa, Francisco and Synnaeve, Gabriel and Usunier, Nicolas and Kirillov, Alexander and Zagoruyko, Sergey},
  booktitle={European conference on computer vision},
  pages={213--229},
  year={2020},
  organization={Springer}
}

@inproceedings{mask2former,
  title={Masked-attention mask transformer for universal image segmentation},
  author={Cheng, Bowen and Misra, Ishan and Schwing, Alexander G and Kirillov, Alexander and Girdhar, Rohit},
  booktitle={Proceedings of the IEEE/CVF conference on computer vision and pattern recognition},
  pages={1290--1299},
  year={2022}
}

@inproceedings{depth_anything,
  title={Depth anything: Unleashing the power of large-scale unlabeled data},
  author={Yang, Lihe and Kang, Bingyi and Huang, Zilong and Xu, Xiaogang and Feng, Jiashi and Zhao, Hengshuang},
  booktitle={Proceedings of the IEEE/CVF conference on computer vision and pattern recognition},
  pages={10371--10381},
  year={2024}
}

@inproceedings{clip,
  title={Learning transferable visual models from natural language supervision},
  author={Radford, Alec and Kim, Jong Wook and Hallacy, Chris and Ramesh, Aditya and Goh, Gabriel and Agarwal, Sandhini and Sastry, Girish and Askell, Amanda and Mishkin, Pamela and Clark, Jack and others},
  booktitle={International conference on machine learning},
  pages={8748--8763},
  year={2021},
  organization={PmLR}
}

@inproceedings{dit,
  title={Scalable diffusion models with transformers},
  author={Peebles, William and Xie, Saining},
  booktitle={Proceedings of the IEEE/CVF international conference on computer vision},
  pages={4195--4205},
  year={2023}
}

@inproceedings{mae,
  title={Masked autoencoders are scalable vision learners},
  author={He, Kaiming and Chen, Xinlei and Xie, Saining and Li, Yanghao and Doll{\'a}r, Piotr and Girshick, Ross},
  booktitle={Proceedings of the IEEE/CVF conference on computer vision and pattern recognition},
  pages={16000--16009},
  year={2022}
}

@article{zhao2025diception,
  title={Diception: A generalist diffusion model for visual perceptual tasks},
  author={Zhao, Canyu and Liu, Mingyu and Zheng, Huanyi and Zhu, Muzhi and Zhao, Zhiyue and Chen, Hao and He, Tong and Shen, Chunhua},
  journal={arXiv preprint arXiv:2502.17157},
  year={2025}
}

@inproceedings{deng2009imagenet,
  title={Imagenet: A large-scale hierarchical image database},
  author={Deng, Jia and Dong, Wei and Socher, Richard and Li, Li-Jia and Li, Kai and Fei-Fei, Li},
  booktitle={2009 IEEE conference on computer vision and pattern recognition},
  pages={248--255},
  year={2009},
  organization={Ieee}
}

@inproceedings{zhou2017scene,
  title={Scene parsing through ade20k dataset},
  author={Zhou, Bolei and Zhao, Hang and Puig, Xavier and Fidler, Sanja and Barriuso, Adela and Torralba, Antonio},
  booktitle={Proceedings of the IEEE conference on computer vision and pattern recognition},
  pages={633--641},
  year={2017}
}

@inproceedings{coco,
  title={Microsoft coco: Common objects in context},
  author={Lin, Tsung-Yi and Maire, Michael and Belongie, Serge and Hays, James and Perona, Pietro and Ramanan, Deva and Doll{\'a}r, Piotr and Zitnick, C Lawrence},
  booktitle={European conference on computer vision},
  pages={740--755},
  year={2014},
  organization={Springer}
}

@misc{gpt4,
      title={GPT-4 Technical Report}, 
      author={OpenAI},
      year={2024},
      eprint={2303.08774},
      archivePrefix={arXiv},
      primaryClass={cs.CL},
      url={https://arxiv.org/abs/2303.08774}, 
}

@misc{ds2025r1,
      title={DeepSeek-R1: Incentivizing Reasoning Capability in LLMs via Reinforcement Learning}, 
      author={DeepSeek-AI},
      year={2025},
      eprint={2501.12948},
      archivePrefix={arXiv},
      primaryClass={cs.CL},
      url={https://arxiv.org/abs/2501.12948}, 
}
\section{Appendix}
\subsection{From Task-Specific Pipelines to Universal Flow}
\label{sec:motivation}

\renewcommand{\dblfloatpagefraction}{.9}
\begin{figure}[!htbp]
    \centering
    \includegraphics[width=0.45\textwidth]{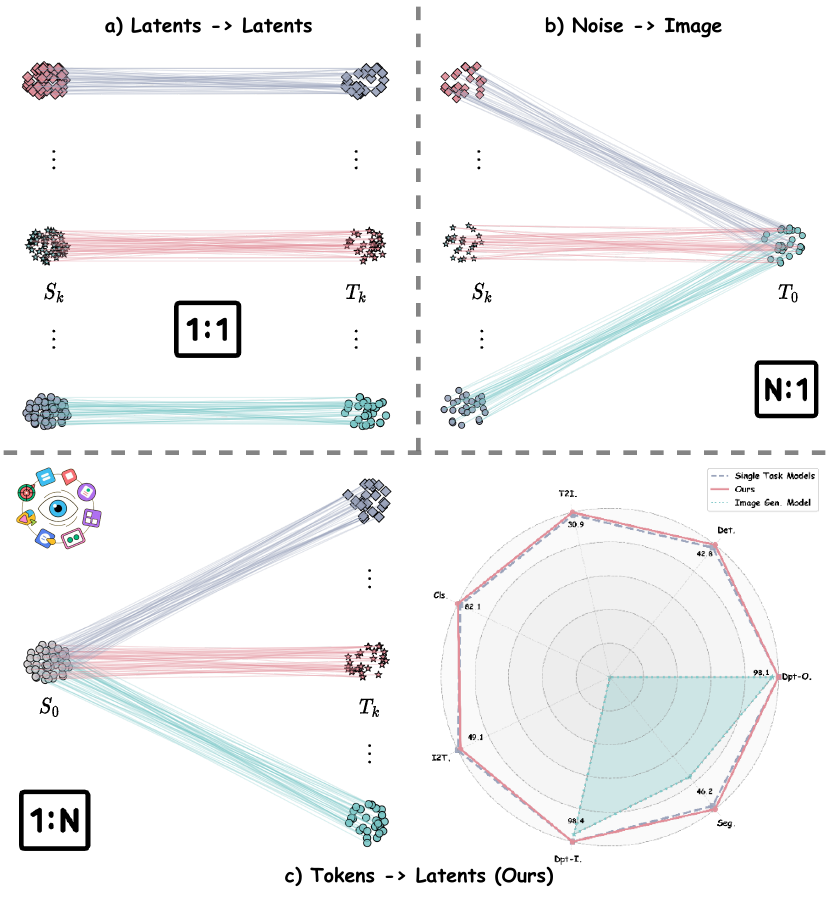}
    \caption{\textbf{Architectural comparison across paradigms for univeral vision modeling.} 
    (a)~Traditional task-specific models adopt a \emph{latent-to-latent} mapping, where each downstream task requires a dedicated architecture and training pipeline, leading to poor scalability and high deployment cost. 
    (b)~Fully generative approaches reconstruct visual outputs by denoising from random noise, treating all tasks as image generation. While flexible for dense prediction, they struggle with non-generative tasks (e.g., classification, retrieval) and often produce semantically inconsistent predictions without strong conditioning. 
    (c)~Our method leverages a self-supervised tokenizer to extract semantic-aware tokens, then employs \emph{flow matching} to model the dynamic transformation from tokens to task-specific latents. This enables precise, structured, and efficient knowledge routing across diverse vision tasks—including generative, discriminative, and metric-based tasks—within a univeral framework.}
    \label{fig:arch_comparison}
\end{figure}

Modern vision systems have evolved from isolated, task-specific models to attempts at unified multi-task architectures. As illustrated in Fig.~\ref{fig:arch_comparison}a, conventional approaches rely on dedicated networks—such as U-Net for segmentation or DPT for depth estimation—that map input features to task-specific outputs through fixed latent spaces. While these models achieve strong performance, their lack of parameter sharing and architectural redundancy hinder scalability in real-world deployment.

A recent trend explores fully generative frameworks that synthesize outputs by reversing a diffusion process from pure noise~(Fig.~\ref{fig:arch_comparison}b)~\cite{zhao2025diception, le2024diffusiongenerate}. These methods unify tasks under a single generative paradigm but face two critical limitations. First, they treat perception as reconstruction, often neglecting input semantics during generation. Without strong conditioning, they are prone to hallucinations and spatial inaccuracies. Second, and more fundamentally, \textbf{they are inherently incompatible with non-generative tasks such as image classification and retrieval}. These tasks require learning discriminative or metric-aware representations, not pixel-level synthesis. Forcing them into a generation pipeline either requires ad-hoc modifications or results in suboptimal performance.

In contrast, our framework introduces a structured and task-agnostic alternative: we decouple \emph{semantic understanding} from \emph{task execution}. First, a self-supervised tokenizer (e.g., DINOv2~\cite{oquab2023dinov2}) extracts compact, input-aware tokens that encode rich contextual and spatial information. These tokens serve as a shared, disentangled representation across all tasks. Then, we formulate task adaptation as a \emph{conditional flow matching} process~\cite{lipman2022flow}, where a velocity field is learned to transform the input tokens into task-specific latent representations.

This design offers three key advantages:
\textbf{(1)}~Predictions are grounded in the original input semantics, avoiding the instability of pure generation;
\textbf{(2)}~The flow-based decoding is flexible: for dense tasks (e.g., segmentation, depth), it generates structured outputs; for discriminative tasks (e.g., classification), the final token representation can be directly pooled and classified; for metric tasks (e.g., retrieval), the token embeddings serve as robust global descriptors;
\textbf{(3)}~Minimal architectural changes are required—only a lightweight flow head or classifier head is added per task.

By bridging discriminative tokenization with generative or discriminative decoding, our method achieves both strong generalization and broad task coverage, establishing a truly univeral paradigm for scalable vision systems.

\subsection{Further ablation on task embedding and decoder architecture.}
To further investigate the impact of task conditioning and latent space design, we evaluate several variants of our framework using different task embeddings and decoders. As shown in Table~\ref{tab:ablation_task_emb}, we compare: (i) CLIP-aligned text embeddings as task prompts (similar to OneDiffusion(\cite{le2024diffusiongenerate})); (ii) randomly initialized task vectors; and (iii) our SD3-based VAE with a $16 \times 32 \times 32$ latent space. While both CLIP-based and random embeddings provide basic task discrimination, they underperform compared to our full model—indicating that semantic priors alone are insufficient for precise task routing in a unified framework. The CLIP text embedding, despite its strong zero-shot generalization in VLMs, suffers from misalignment when directly used to condition latent dynamics~\cite{zhao2025diception}, likely due to domain gap between natural language and visual token flows. Random embeddings, though straight-forward, lack semantic structure and lead to unstable optimization. This is reflected in the jitter of the Loss curve during convergence and the lack of final performance.
\begin{table}[!htbp]
\centering
\caption{Ablation study on task embedding and decoder design.}
\label{tab:ablation_task_emb}
\begin{tabular}{lcc}
\toprule
\textbf{Method} & \textbf{ImageNet-1K} \\
\midrule
Ours (CLIP text as task\_emb) & 80.6  \\
Ours (Random task\_emb)       & 81.3 \\
Ours (SD3 VAE encoder)        & 79.5   \\
\midrule
DeiT-B Baseline & \textbf{81.8}  \\
\midrule
\textcolor{graytext}{Ours (Zero-shot)} & \underline{\textcolor{graytext}{81.5}} \\
\bottomrule
\end{tabular}
\end{table}

Furthermore, replacing our flow-based encoder with a standard SD3 VAE—despite its powerful generative capacity—results in degraded performance (\texttt{Acc: 79.5\%}. This is because the VAE is optimized for image reconstruction, not structured latent transformation, and fails to preserve fine-grained spatial coherence required for dense tasks. In contrast, our conditional flow matching operates directly in the compact token space with visual priors and learns a dynamic path from input-aware tokens to task-specific latents, enabling both high fidelity and strong generalization.
\subsection{Analysis of Segmentation Results}
\label{sec:appendix_seg}

\begin{figure}[!htpb]
    \centering
    \includegraphics[width=0.45\textwidth]{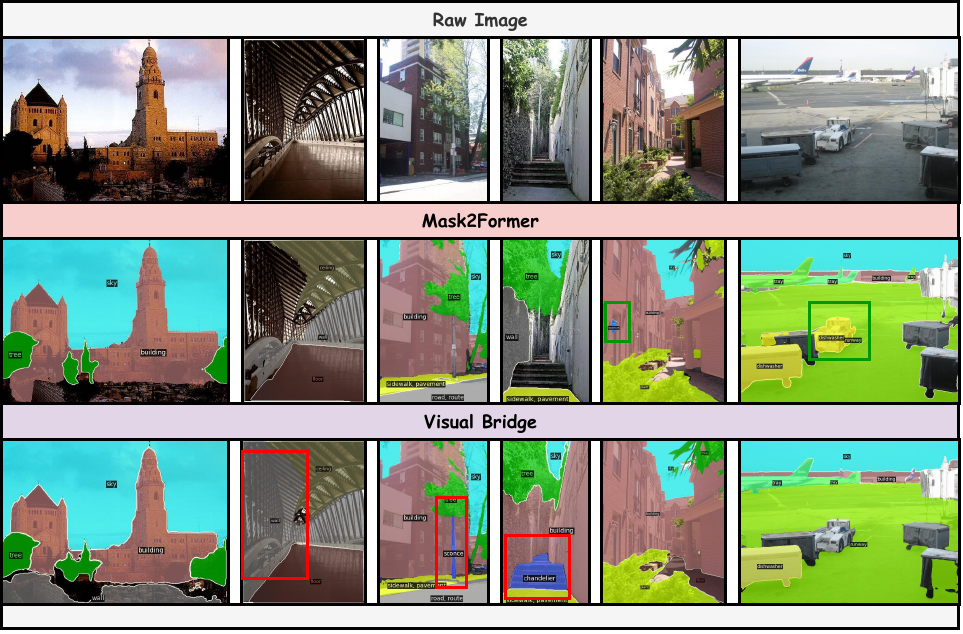}
    \caption{\textbf{Zero-shot segmentation results on ADE20K.} Despite being trained in a generative manner, our method produces accurate and coherent segmentations. In challenging scenes involving fine-grained object boundaries (e.g., sconces), occlusions, or complex layouts, our model sometimes outperforms the original Mask2Former baseline~\cite{mask2former}, demonstrating stronger contextual understanding and boundary coherence. We attribute this improvement to the low-variance latent prediction in our generative framework, which encourages global scene consistency and robust feature abstraction.}
    \label{fig:seg_result}
\end{figure}

We present qualitative comparisons of zero-shot segmentation results in Figure~\ref{fig:seg_result}, evaluating our method on the ADE20K without any task-specific fine-tuning. Although our model is trained through a generative objective—predicting image latents in a diffusion-like framework—it achieves competitive, and in many cases superior, segmentation accuracy compared to the original discriminative Mask2Former~\cite{mask2former}.

Notably, our approach excels in scenes with ambiguous boundaries, object occlusions, and complex spatial layouts. For instance, in images containing tightly packed furniture or overlapping instances, our model produces more coherent and semantically plausible masks, while the baseline tends to fragment objects or misalign boundaries. This suggests that our method captures stronger global context and structural priors during inference.

We attribute this enhanced performance to the \textit{low-variance latent prediction} inherent in our generative formulation. Unlike discriminative models that optimize for per-pixel classification, our framework learns to reconstruct the entire scene in a compressed latent space, which implicitly regularizes the representation and encourages holistic understanding. This leads to more stable and globally consistent predictions, especially in ambiguous or low-contrast regions.

The results highlight a key advantage of generative knowledge integration: by learning to \textit{synthesize} rather than merely \textit{classify}, the model acquires richer internal representations that generalize well under zero-shot conditions. This differentiates our approach from conventional task-specific models and underscores the potential of generative frameworks in building univeral, generalist vision systems.

\subsection{Analysis of Depth Estimation Performance}
\label{sec:appendix_depth}

\begin{figure}[!htpb]
    \centering
    \includegraphics[width=0.45\textwidth]{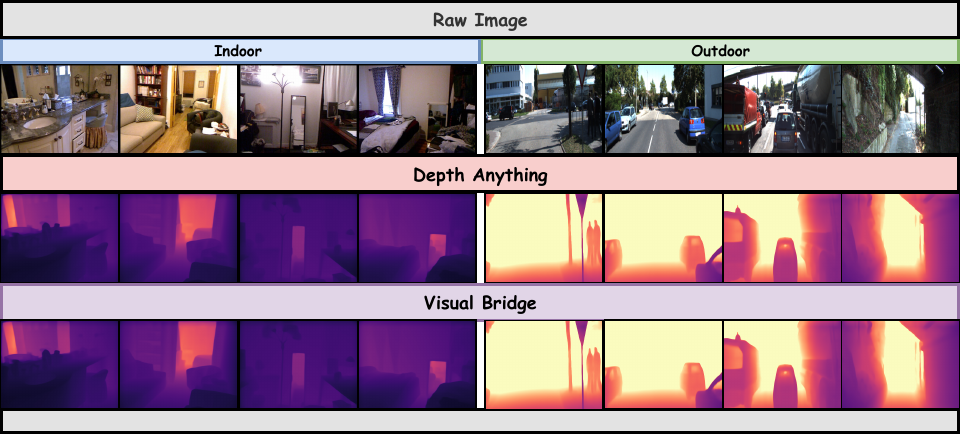}
    \caption{\textbf{Qualitative comparison of zero-shot monocular depth estimation.} Our method produces depth maps that are comparable in quality to those of Depth Anything~\cite{depth_anything}, capturing fine-grained details such as object boundaries, surface normals, and large-scale scene geometry. Despite being trained within a univeral generative framework alongside diverse vision tasks, our model achieves on-par performance without task-specific architectural designs or post-processing. This demonstrates the strong generalization capability of our approach in dense prediction tasks.}
    \label{fig:depth_result}
\end{figure}

Quantitatively, our model achieves comparable performance on standard metrics (e.g., Abs Rel, $\delta$) benchmarks under zero-shot evaluation, despite not being explicitly optimized for depth prediction.
We evaluate the depth estimation capability of our model in a zero-shot setting and compare it qualitatively with Depth Anything~\cite{depth_anything}, a state-of-the-art dedicated monocular depth estimator. As shown in Figure~\ref{fig:depth_result}, our method produces depth maps that are visually on par with Depth Anything across diverse scenes—including indoor environments, urban landscapes, and natural terrains. Fine details such as depth discontinuities at object edges, relative depth between overlapping instances, and continuous surface gradients are well preserved.

The strong performance in depth estimation further validates the effectiveness of our univeral framework. By learning to reconstruct scene latents through a generative objective, the model implicitly captures geometric priors and multi-scale contextual cues necessary for dense prediction tasks.

\end{document}